\documentclass{article}
\usepackage{ijcai22}

\usepackage{times}
\usepackage{soul}
\usepackage{url}
\usepackage[hidelinks]{hyperref}
\usepackage[utf8]{inputenc}
\usepackage[small]{caption}
\usepackage{graphicx}
\usepackage{amsmath}
\usepackage{amsthm}
\usepackage{booktabs}
\usepackage{algorithm}
\usepackage{algorithmic}
\usepackage{multirow}
\title{Curriculum-Based Self-Training Makes Better Few-Shot Learners for Data-to-Text Generation}

\author{
    Pei Ke$^1$\and Haozhe Ji$^1$\and Zhenyu Yang$^2$\and Yi Huang$^{3,4}$\and Junlan Feng$^{3,4}$\and \\ 
    Xiaoyan Zhu$^1$\And Minlie Huang$^{1,4}$\thanks{Corresponding author} \\
    \affiliations
    $^1$CoAI Group, DCST, IAI, BNRIST, Tsinghua University, Beijing, China \\
    $^2$OPPO Mobile Telecommunications Corp., Ltd, China \\
    $^3$JIUTIAN Team, China Mobile Research Institute, Beijing 100053, China \\
    $^4$Tsinghua University-China Mobile Communications Group Co., Ltd. Joint Institute, Beijing, China \\
    \emails
    \{kp17, jhz20\}@mails.tsinghua.edu.cn, yangzhenyu@oppo.com \\ \{huangyi, fengjunlan\}@chinamobile.com, \{zxy-dcs, aihuang\}@tsinghua.edu.cn
}

\begin{document}

\maketitle

\begin{abstract}

Despite the success of text-to-text pre-trained models in various natural language generation (NLG) tasks, the generation performance is largely restricted by the number of labeled data in downstream tasks, particularly in data-to-text generation tasks. Existing works mostly utilize abundant unlabeled structured data to conduct unsupervised pre-training for task adaption, which fail to model the complex relationship between source structured data and target texts. Thus, we introduce self-training as a better few-shot learner than task-adaptive pre-training, which explicitly captures this relationship via pseudo-labeled data generated by the pre-trained model. To alleviate the side-effect of low-quality pseudo-labeled data during self-training, we propose a novel method called Curriculum-Based Self-Training (CBST) to effectively leverage unlabeled data in a rearranged order determined by the difficulty of text generation. Experimental results show that our method can outperform fine-tuning and task-adaptive pre-training methods, and achieve state-of-the-art performance in the few-shot setting of data-to-text generation.

\end{abstract}

\section{Introduction}

Recently, text-to-text pre-trained models like BART \cite{lewis2020bart} and T5 \cite{raffel2020t5} have emerged in the field of natural language generation (NLG). Their main idea is to capture the relationship among texts by reconstructing original sentences with the input of corrupted sentences. These models can achieve state-of-the-art performance in various NLG tasks via fine-tuning on downstream datasets.

\begin{figure}[!htp]
  \centering
  \includegraphics[width=1.0\linewidth]{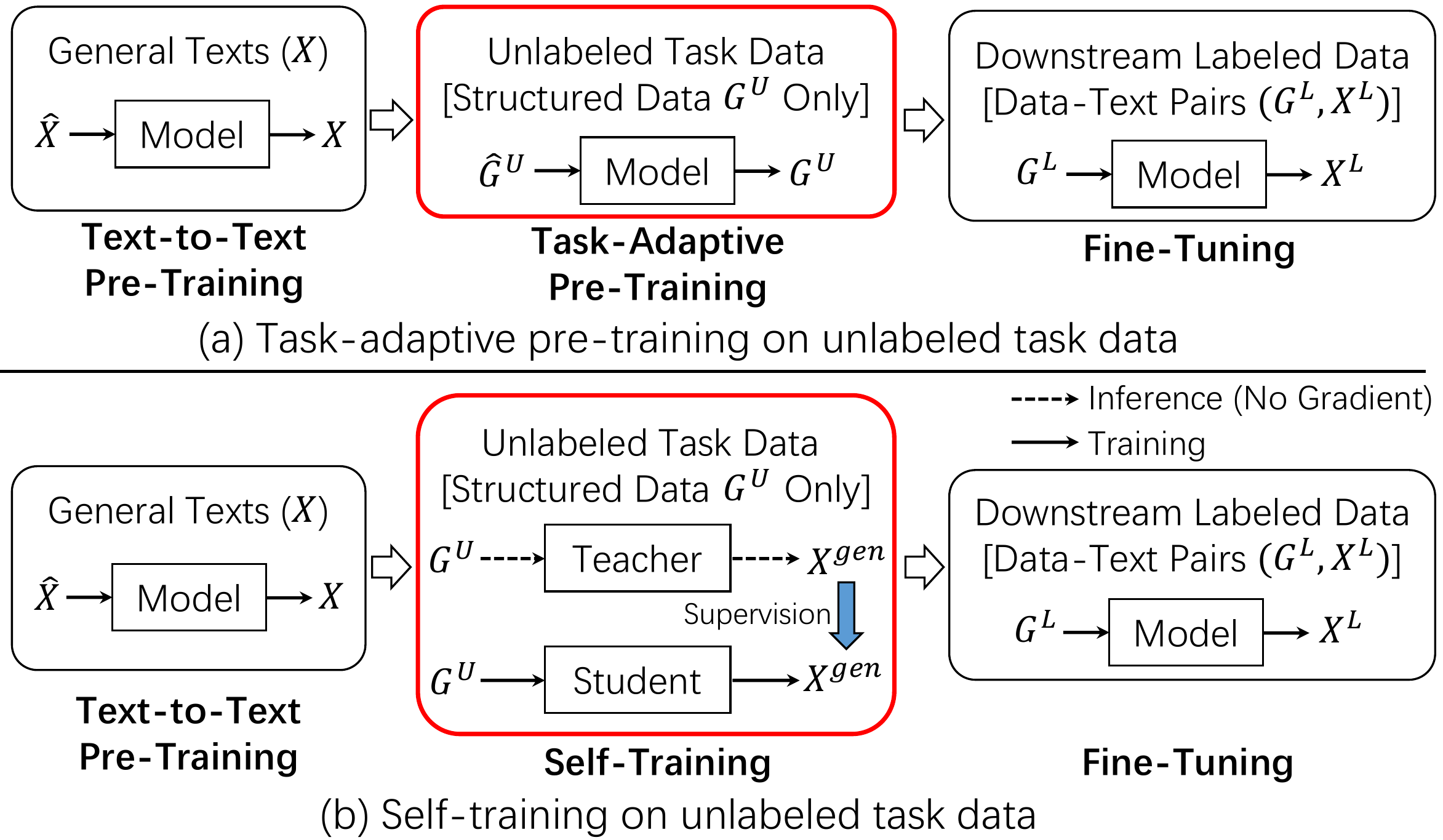}% 1\linewidth
  \caption{Comparison between task-adaptive pre-training and self-training in few-shot data-to-text generation. Task-adaptive pre-training adopts unsupervised pre-training methods on the unlabeled task data $G^U$ (such as structured data reconstruction), while self-training instead generates texts $X^{gen}$ for $G^U$ with the teacher model and trains the student model on the generated pseudo-labeled data.
  $\hat{G}^U$ / $\hat{X}$ denotes the corrupted structured data / texts.}
  \label{fig:intro}
\end{figure}

Despite the success of text-to-text pre-trained models, it's challenging to directly apply them to few-shot NLG tasks because their performance can be largely restricted by the number of labeled data \cite{chen2020kgpt,chen2020fewshotnlg}. Especially in the task of few-shot data-to-text generation, it's hard to learn the complex relationship between source structured data and target texts via limited labeled data \cite{gong2020tablegpt}.
Existing works commonly continue pre-training on large amounts of structured data without corresponding texts for task adaption, aiming to enhance the model performance \cite{guru2020dontstop}.
As shown in Figure \ref{fig:intro}(a), this task-adaptive pre-training strategy appears between general pre-training and fine-tuning, whose purpose is to adapt general pre-trained models (such as BART) to specific tasks (such as data-to-text generation) using unlabeled task data \cite{xing2021structurepretrain}.
We argue that task-adaptive pre-training only captures the relationship among structured data but fails to explicitly model the alignment between structured data and texts, thereby restricting the model performance on the few-shot data-to-text generation tasks.

In this paper, we introduce self-training \cite{scudder1965selftraining} as a better few-shot learner for data-to-text generation. As shown in Figure \ref{fig:intro}(b), self-training is a teacher-student framework where the teacher model creates synthetic labels for the unlabeled data, and the student model is trained on the pseudo-labeled data constructed by the teacher model.
Compared with task-adaptive pre-training, self-training explicitly models the relationship between structured data and texts by training the student model on the pseudo-labeled data generated by the teacher model, instead of solely conducting unsupervised pre-training on the unlabeled structured data.
We argue that the main challenge falls into the quality of pseudo-labeled data. Even if the teacher model is initialized with pre-trained models, it may generate low-quality texts especially when dealing with the structured data with complex topologies, which hurt the performance of the student model.

Thus, we present a novel method called Curriculum-Based Self-Training (CBST) to address the challenge. This method utilizes curriculum learning \cite{bengio2009curriculum} to construct pseudo-labeled data from easy cases to hard ones, and leverages such data into the training process at different iterations.
Specifically, we divide the unlabeled dataset into different subsets based on some difficulty metric such as the number of input triples \cite{ribeiro2020globallocal}. At each iteration, we first collect the unlabeled data that satisfy the difficulty metric of the current curriculum. Then, we generate synthetic texts for these data using the teacher model, and select the pseudo-labeled data based on the coverage and generation probability. Finally, we train the student model on the labeled data and the selected pseudo-labeled data, and make it act as the teacher model at the next iteration. This method is expected to gradually increase the difficulty of generating texts for unlabeled structured data and improve the quality of pseudo-labeled data constructed by the teacher model.

Our contributions are mainly as follows\footnote{The codes are available at \url{https://github.com/kepei1106/CBST}.}:
\begin{itemize}
    \item We introduce self-training to improve text-to-text pre-training for few-shot data-to-text generation. This method explicitly captures the relationship between structured data and texts by generating texts for unlabeled structured data with the teacher model and training the student model on the pseudo-labeled data.
    
    \item We propose a novel method called CBST to alleviate the side-effect of low-quality pseudo-labeled data, which introduces curriculum learning to effectively leverage the unlabeled structured data into the training process in the order determined by the difficulty metric.

    \item We conduct extensive experiments in the few-shot setting of WebNLG and WikiBio datasets. Results show that our method can outperform fine-tuning and task-adaptive pre-training methods, and achieve state-of-the-art performance on these two benchmarks.
\end{itemize}

\begin{figure*}[!htp]
  \centering
  \includegraphics[width=0.95\linewidth]{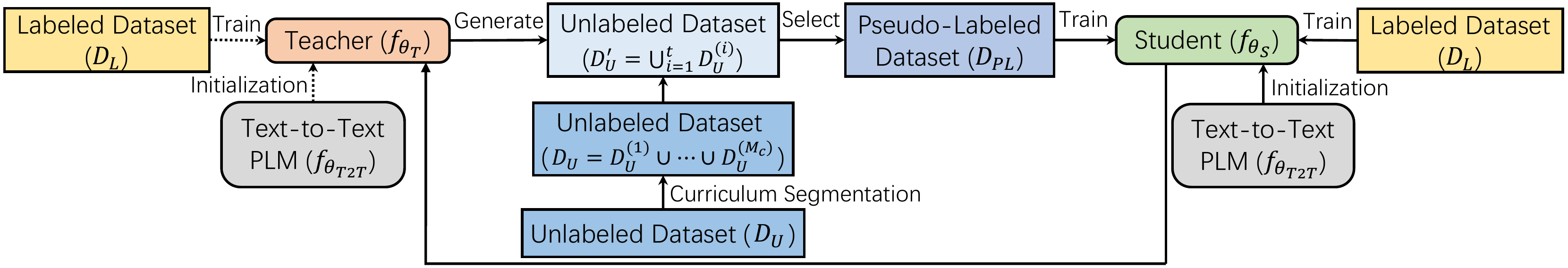}% 1\linewidth
  \caption{Overview of CBST. Dotted lines indicate that the teacher is initialized with the text-to-text pre-trained model and trained on the labeled dataset only before the first iteration. Afterward, the teacher model is initialized by the student model from the previous iteration. $M_C$ denotes the number of curriculum, and $D_U^{'}$ contains the unlabeled data included in the current curriculum at each iteration $t$.}
  \label{fig:overview}
\end{figure*}

\section{Related Work}

\subsection{Data-to-Text Generation}

Early studies on data-to-text generation mainly focus on how to encode the structural information of input data better. Thus, researchers devise complex encoder structures based on sequential neural networks \cite{bayu2018gtrlstm} and graph neural networks \cite{ribeiro2020globallocal} to achieve this goal.
Recently, since text-to-text pre-trained models have shown promising performance in various NLG tasks, some works directly fine-tune these pre-trained models including BART \cite{lewis2020bart} and T5 \cite{raffel2020t5} on data-to-text datasets and report impressive results \cite{ribeiro2020investigate,kale2020text}. Other works adapt pre-trained models to data-to-text generation tasks by designing specific pre-training tasks such as reconstructing structured data and texts, which further improve the model performance in both supervised and few-shot settings of downstream datasets \cite{chen2020kgpt,ke2021jointgt,xing2021structurepretrain}.

Compared with existing works, we introduce self-training to explicitly capture the relationship between structured data and texts, instead of solely relying on unsupervised task-adaptive pre-training. Our method is expected to utilize unlabeled data more effectively and further improve text-to-text pre-trained models in few-shot data-to-text generation.

\subsection{Self-Training}

Self-training \cite{scudder1965selftraining} is a teacher-student framework to leverage unlabeled data through semi-supervised learning.
Self-training has been applied to the tasks of text classification \cite{du2021selftraining} and generation \cite{he2020revisit}.
With the development of pre-trained models, recent works also show that self-training is complementary to pre-training for various classification tasks \cite{du2021selftraining}.

For comparison, our work is the first attempt to utilize self-training to improve text-to-text pre-trained models for few-shot data-to-text generation. Moreover, we consider the difficulty of generating texts for different structured data and introduce curriculum learning \cite{bengio2009curriculum} to incorporate unlabeled data into the self-training process at different iterations to improve the quality of pseudo-labeled data.

\begin{figure}[!t]
  \centering
  \includegraphics[width=1.0\linewidth]{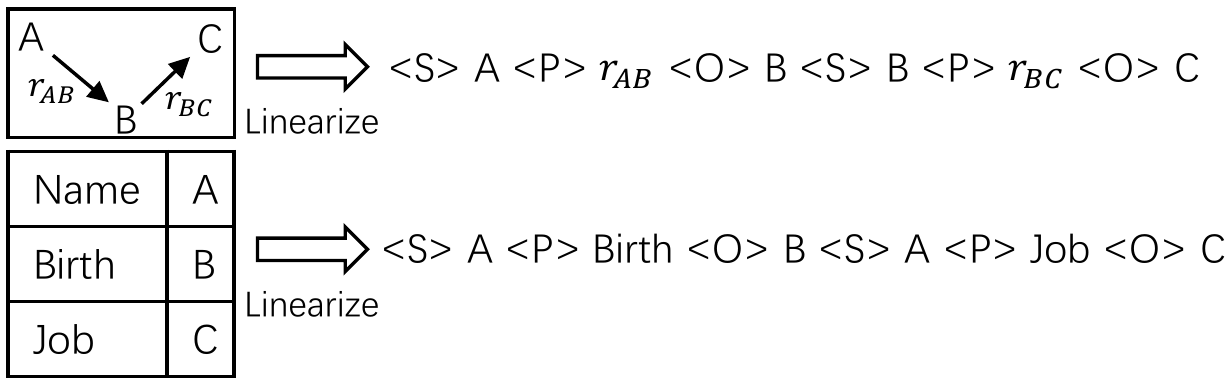}% 1\linewidth
  \caption{Linearization of structured data including graphs and tables. $<$S$>$ / $<$P$>$ / $<$O$>$ is the special token indicating subjects / predicates / objects, respectively.}
  \label{fig:linear}
\end{figure}

\section{Method}

\subsection{Task Definition and Model Overview}

Given the text-to-text pre-trained model $f_{\theta_{T2T}}$, the downstream labeled dataset $D_L=\left\{(G_i^L,X_i^L)\right\}_{i=1}^m$, and the unlabeled dataset $D_U=\{G_i^U\}_{i=1}^M$ where $G_i^L$ / $G_i^U$ denotes the structured data and $X_i^L$ indicates the annotated texts ($M\gg m$), our goal is to obtain a data-to-text pre-trained model which can perform well in the few-shot setting of downstream datasets.

The overview of our proposed method is shown in Figure \ref{fig:overview}.
We follow the existing works to linearize the structured data as a sequence of triples containing subjects, predicates, and objects \cite{ribeiro2020investigate,chen2020kgpt} as shown in Figure \ref{fig:linear}.
We first initialize the teacher model $f_{\theta_{T}}$ with the pre-trained model $f_{\theta_{T2T}}$ and train it on the labeled dataset $D_L$. Then, we train the teacher model and the student model iteratively. 
At each iteration, the teacher model $f_{\theta_{T}}$ generates synthetic texts for the unlabeled dataset $D_U^{'}$ which contains the data satisfying the difficulty metric of the current curriculum.
Then we select the pseudo-labeled dataset $D_{PL}$ based on the coverage and generation probability of the teacher model. The student model $f_{\theta_{S}}$ which is also initialized with the pre-trained model $f_{\theta_{T2T}}$ is trained on the pseudo-labeled dataset $D_{PL}$ and the labeled dataset $D_L$. Finally, the student model $f_{\theta_{S}}$ acts as the teacher model $f_{\theta_{T}}$ in the next iteration.

\subsection{Curriculum-Based Self-Training}

Our proposed method is shown in Algorithm \ref{alg:clest}, which consists of three main steps: curriculum segmentation, pseudo-labeled data construction, and student model training.

\subsubsection{Curriculum Segmentation}
\label{sec:curseg}

Curriculum segmentation is designed based on the difficulty of generating texts for structured data. In this paper, we choose the number of triples as the difficulty metric to segment the unlabeled dataset $D_U$ into $M_C$ subsets. This metric reflects the complexity of structures in the input data and has a significant impact on the quality of the generated texts \cite{ribeiro2020globallocal}. We also try other metrics in \S \ref{sec:ablation}.

\subsubsection{Pseudo-Labeled Data Construction}
\label{sec:pseudodata}

This module aims to generate pseudo-labeled data with the teacher model at each iteration.
Prior to the first iteration, the teacher model $f_{\theta_T}$ is initialized with the pre-trained model $f_{\theta_{T2T}}$ and trained on the labeled dataset $D_L$ with the following loss function:
\begin{equation}
\label{eqn:init}
    \mathcal{L}_{init} = \frac{1}{m}\sum_{i=1}^m l(X_i^L,G_i^L;f_{\theta_T})
\end{equation}
where $l(X,G;f_\theta)$ indicates the negative log-likelihood loss and is commonly used in the sequence generation tasks:
\begin{equation}
    l(X,G;f_\theta) = -\sum_{t=1}^{|X|} \log P_{\theta}(X_t|G,X_{<t})
\end{equation}
Note that except for the first iteration, the teacher model is initialized by the student model from the previous iteration. At each iteration, we first utilize the teacher model to generate the synthetic texts for the unlabeled dataset $D_U^{'}\subset D_U$ which includes the unlabeled structured data that satisfy the difficulty metric of the current curriculum:
\begin{equation}
\label{eqn:teachergen}
    X_i^{gen} = f_{\theta_T}(G_i^U), i=1,2,\cdots,|D_U^{'}|
\end{equation}

Then, we select the generated texts based on coverage and generation probability which reflect the generation quality \cite{gehrmann2018coverage} to construct the pseudo-labeled dataset $D_{PL}$ with $M^{'}$ samples.
Specifically, we choose the pseudo-labeled data where the proportion of subjects and objects appearing in the generated text is larger than $\epsilon_{cov}$ and the generation probability of the text ranks top-$\epsilon_{gen}$ to construct the dataset for the student model training. Here, $\epsilon_{cov}$ and $\epsilon_{gen}$ are both hyper-parameters.

\begin{algorithm}[!t] 
\caption{Curriculum-Based Self-Training (CBST)}
\label{alg:clest} 
\begin{algorithmic}[1]
\REQUIRE ~~\\
Labeled dataset: $D_L=\{(G_i^L,X_i^L)\}_{i=1}^m$  \\
Unlabeled dataset: $D_U=\{G_i^U\}_{i=1}^M$ \\
Curriculum segmentation criterion: $C$ \\
Text-to-text pre-trained model: $f_{\theta_{T2T}}$ \\
\ENSURE ~~\\
Data-to-text pre-trained model: $f_{\theta_{D2T}}$ \\
\STATE Use $C$ to split $D_U$ into $D_U^{(1)},D_U^{(2)},...,D_U^{(M_C)}$
\STATE Initialize the teacher model $f_{\theta_T}$ with $f_{\theta_{T2T}}$ and train $f_{\theta_T}$ on $D_L$ via Eq.(\ref{eqn:init})
%\STATE $D_N^{(0)}=\emptyset$
\FOR{$t=1,2,\cdots,M_C$}
\STATE $D_U^{'}\leftarrow \bigcup_{i=1}^t D_U^{(i)}$
\STATE Apply $f_{\theta_T}$ to the unlabeled dataset $D_U^{'}$ to acquire the generated texts via Eq.(\ref{eqn:teachergen})
\STATE Select the pseudo-labeled data based on coverage and generation probability to construct the dataset $D_{PL}$ 
\STATE Initialize the student model $f_{\theta_S}$ with $f_{\theta_{T2T}}$ and train $f_{\theta_S}$ on $D_{PL}\cup D_L$ with input noise via Eq.(\ref{eqn:student})
\STATE $f_{\theta_T}\leftarrow f_{\theta_S}$
\ENDFOR
\STATE $f_{\theta_{D2T}}\leftarrow f_{\theta_S}$
\RETURN $f_{\theta_{D2T}}$
\end{algorithmic}
\end{algorithm}

\subsubsection{Student Model Training}
\label{sec:studenttrain}

The goal of this module is to train the student model on the pseudo-labeled dataset and the labeled dataset. Initialized with the text-to-text pre-trained model $f_{\theta_{T2T}}$, the student model $f_{\theta_S}$ is trained with the following loss function:
\begin{align}
\label{eqn:student}
    \mathcal{L}_{S}=\frac{1}{M^{'}}&\sum_{i=1}^{M^{'}} l(X_i^{gen},\textrm{NF}(G_i^U);f_{\theta_S}) \notag \\
    & + \frac{1}{m}\sum_{i=1}^m l(X_i^L,G_i^L;f_{\theta_S})
\end{align}
where $\textrm{NF}(G)$ denotes the noise function perturbing the structured data $G$. Existing works show that input noise is beneficial for self-training because it increases the smoothness of the student model \cite{he2020revisit}. Inspired by the existing works \cite{jia2019substitution,hoyle2020promoting}, we devise two-level noise functions as shown in Table \ref{tab:noise}, including word substitution at the semantic level and triple reordering at the structural level. Specifically, we substitute each word by its synonym with the probability $p_{word}$ and shuffle the order of triples in each sample with the probability $p_{triple}$, where $p_{word}$ and $p_{triple}$ are hyper-parameters. These functions are expected to perturb the input structured data while largely keeping the semantic and structural information.

\begin{table} [!t]
\centering
\small
\setlength{\tabcolsep}{1.0mm}{
\begin{tabular}{l}
\hline
Word Substitution: Replace words with their synonyms \\
\hline
Example Input:  \\
$<$S$>$ Bakewell \textbf{pudding} $<$P$>$ \textbf{served} $<$O$>$ Warm or cold \\
\hline
Example Output: \\
$<$S$>$ Bakewell \textbf{dessert} $<$P$>$ \textbf{acted} $<$O$>$ Warm or cold \\
\hline
\hline
Triple Reordering: Shuffle the order of triples \\
\hline
Example Input:  \\
$<$S$>$ Alan Shepard $<$P$>$ status $<$O$>$ Deceased \\
$<$S$>$ Alan Shepard $<$P$>$ occupation $<$O$>$ Test pilot \\
\hline
Example Output: \\
$<$S$>$ Alan Shepard $<$P$>$ occupation $<$O$>$ Test pilot \\
$<$S$>$ Alan Shepard $<$P$>$ status $<$O$>$ Deceased \\
\hline
\end{tabular}}
\caption{Noise functions. $<$S$>$ / $<$P$>$ / $<$O$>$ is the special token indicating the subject / predicate / object of the input triples, respectively. Bold texts indicate the words for substitution.}
\label{tab:noise}
\end{table}

In practice, we adopt a separate training strategy which shows better performance than joint training \cite{he2020revisit}. This strategy first trains the student model on the pseudo-labeled dataset $D_{PL}$ with the first term of $\mathcal{L}_{S}$, and then trains it on the labeled dataset $D_L$ with the second term.

\section{Experiment}

\subsection{Dataset}

\paragraph{WebNLG.} This dataset aims to generate textual descriptions for RDF triples \cite{shimorina2018webnlgv2}. The number of instances in training / validation / test set is 34,352 / 4,316 / 4,224, respectively. We followed the existing works \cite{chen2020kgpt} to pre-process this dataset and use 0.5\%, 1\%, 5\%, 10\% of the training instances as the labeled datasets in the few-shot setting.

\paragraph{WikiBio.} This dataset aims to generate the first sentence of biography descriptions for Wikipedia tables \cite{lebret2016wikibio}. The original split of the training / validation / test set is 582,658 / 2,000 / 72,831. We followed the existing works \cite{chen2020kgpt,chen2020fewshotnlg} to pre-process this dataset and use 50, 100, 200, and 500 samples from the training dataset as the labeled datasets in the few-shot setting.

We further constructed the unlabeled dataset for each benchmark dataset based on GenWiki \cite{jin2020genwiki}. This dataset consists of general-domain unpaired structured data and texts sharing the same content distribution. We directly removed the texts in this dataset, and filtered out the structured data that do not have the subjects and objects appearing in the corresponding labeled dataset. Thus, we obtained two unlabeled datasets for WebNLG and WikiBio, respectively, which contain 375,408 / 163,804 samples of structured data without annotated texts\footnote{We have conducted triple-level matching on the filtered GenWiki and the test set of WebNLG / WikiBio. The results show that there is no overlap between them.}.

\begin{table*} [!htp]
\centering
\scriptsize
\setlength{\tabcolsep}{1.0mm}{
\begin{tabular}{l|ccc|ccc|ccc|ccc}
\hline
\% of Training Data & \multicolumn{3}{c|}{0.5\%} & \multicolumn{3}{c|}{1\%} & \multicolumn{3}{c|}{5\%} & \multicolumn{3}{c}{10\%} \\
\hline
Model & B-4 & M & R-L & B-4 & M & R-L & B-4 & M & R-L & B-4 & M & R-L \\
\hline
\multicolumn{13}{c}{\textit{Direct Fine-Tuning}}  \\
\hline
FT-KGPT & 22.30 & - & - & 25.60 & - & - & 41.20 & - & - & 47.90 & - & - \\
FT-BART & 31.29 & 29.55 & 53.83 & 38.97 & 33.48 & 58.73 & 51.32 & 40.56 & 66.80 & 55.68 & 42.59 & 69.44 \\
\hline
\multicolumn{13}{c}{\textit{Task-Adaptive Pre-Training}}  \\
\hline
TAPT-SeqRecon & 33.37 & 30.64 & 55.37 & 38.83 & 33.71 & 59.21 & 51.51 & 40.53 & 67.29 & 56.18 & 42.60 & 70.23 \\
TAPT-GraphRecon & 37.47 & 33.31 & 58.84 & 42.29 & 35.93 & 61.22 & 52.25 & 40.78 & 67.96 & 56.57 & 43.02 & 70.44 \\
\hline
\multicolumn{13}{c}{\textit{Self-Training}}  \\
\hline
CBST (Ours) & \textbf{38.74}** & \textbf{34.24}** & \textbf{60.28}** & \textbf{44.40}** & \textbf{37.37}** & \textbf{63.62}** & \textbf{54.98}** & \textbf{42.18}** & \textbf{69.44}** & \textbf{58.78}** & \textbf{43.94}** & \textbf{71.30}** \\
\quad w/o CL & 37.69 & 34.22 & 59.31 & 43.60 & 37.20 & 62.64 & 53.70 & 41.86 & 68.30 & 57.79 & 43.60 & 70.76 \\
\quad w/o CL \& Noise & 37.34 & 34.20 & 58.84 & 43.33 & 37.16 & 62.55 & 52.82 & 41.66 & 67.85 & 56.78 & 43.30 & 70.22 \\
\hline
\end{tabular}}
\caption{BLEU-4(B-4), METEOR(M) and ROUGE-L(R-L) in the different settings of WebNLG. The results of FT-KGPT are re-printed from the original paper of KGPT. - means that the results are not reported in the corresponding reference. ** indicates that our model significantly outperforms the best baseline in the corresponding setting (t-test, $p<0.01$).}
\label{tab:fewshotwebnlg}
\end{table*}

\subsection{Implementation Detail}

In our self-training algorithm, we set the number of curriculum $M_C$ to be 3. For both WebNLG and WikiBio datasets, we split the corresponding unlabeled dataset into 3 subsets which contain the structured data with ($\leq$2) / 3-4 / ($\geq$5) triples. For the hyper-parameters to select pseudo-labeled data, we set $\epsilon_{cov}=1.0,\epsilon_{gen}=50\%$.
The probabilities of word substitution and triple reordering were set to $p_{word}=p_{triple}=0.4$.

As for the model structure, 
we used BART \cite{lewis2020bart} as the text-to-text pre-trained model in our experiments.
The base version of BART was adopted because of the limited computational resources. 
We followed BART to use Byte-Pair Encoding vocabulary with the size of 50,265.
The training epoch at each iteration was set to be 20. 
The learning rate was 0.00003. The batch size was 32 / 24 for WebNLG / WikiBio, respectively. The maximum length of linearized structured data was 256 / 384 for WebNLG / WikiBio, respectively, while the length of text sequences was 128.

\subsection{Baseline}

\paragraph{Direct Fine-Tuning.} This category of baselines directly fine-tunes the state-of-the-art pre-trained models including KGPT \cite{chen2020kgpt}, Switch-GPT-2 \cite{chen2020fewshotnlg} and BART \cite{lewis2020bart} on the labeled data without the use of unlabeled data. We denoted these baselines as FT-KGPT, FT-Switch-GPT-2 and FT-BART, respectively.

\paragraph{Task-Adaptive Pre-Training.} This category of baselines designs task-adaptive pre-training methods on unlabeled data before fine-tuning. We chose two representative pre-training tasks: 1) Sequence-level reconstruction (SeqRecon) from BART \cite{lewis2020bart} which decodes complete linearized structured data when encoding corrupted structured data;
2) Graph-level reconstruction (GraphRecon) from JointGT \cite{ke2021jointgt} which predicts the masked tokens at the output layer of the encoder with the input of corrupted structured data.
We denoted them as TAPT-SeqRecon and TAPT-GraphRecon, respectively.

For a fair comparison, we also used BART \cite{lewis2020bart} as the text-to-text pre-trained model for task-adaptive pre-training baselines.
We presented the results of our model \textbf{CBST} with two ablation models, i.e., \textbf{CBST w/o CL} and \textbf{CBST w/o CL \& Noise}. The former ablation model removes curriculum learning from CBST, which is similar to noisy self-training \cite{he2020revisit}. The latter one simultaneously removes curriculum learning and input noise, which acts as vanilla self-training.
All the results were presented with the average values over 3 runs.

\subsection{Automatic Result}
\label{sec:automatic}

\begin{table} [!t]
\centering
\scriptsize
\setlength{\tabcolsep}{1.0mm}{
\begin{tabular}{l|c|c|c|c}
\hline
\# of Training Data & 50 & 100 & 200 & 500 \\
\hline
 \multicolumn{5}{c}{\textit{Direct Fine-Tuning}}  \\
 \hline
FT-Switch-GPT-2 & 17.20 & 23.80 & 25.40 & 28.60 \\
FT-KGPT & 24.20 & 27.60 & 29.10 & 30.00 \\
 FT-BART & 25.83 & 29.38 & 32.53 & 34.80 \\
 \hline
 \multicolumn{5}{c}{\textit{Task-Adaptive Pre-Training}}  \\
 \hline
TAPT-SeqRecon & 27.93 & 30.81 & 32.59 & 34.90  \\
TAPT-GraphRecon & 28.86 & 30.40 & 32.49 & 35.24 \\
 \hline
 \multicolumn{5}{c}{\textit{Self-Training}}  \\
 \hline
 CBST (Ours) & \textbf{29.63}* & \textbf{31.23}* & \textbf{33.21}* & \textbf{36.27}** \\
 \quad w/o CL  & 29.05 & 30.70 & 33.13 & 35.03 \\
 \quad w/o CL \& Noise & 27.96 & 30.54 & 32.65 & 34.87 \\
\hline
\end{tabular}}
\caption{BLEU-4 in the different settings of WikiBio. The results of FT-Switch-GPT-2 and FT-KGPT are re-printed from the original paper of KGPT. * indicates that our model significantly outperforms the best baseline in the
corresponding setting (t-test, $p<0.05$), while ** means $p<0.01$.}
\label{tab:fewshotwikibio}
\end{table}

We followed the existing works \cite{shimorina2018webnlgv2,chen2020kgpt} to adopt BLEU-4 \cite{papineni2002bleu}, METEOR \cite{banerjee2005meteor}, and ROUGE-L \cite{lin2004rouge} to evaluate the generated results on WebNLG, and use BLEU-4 as the metric on WikiBio. 

The main results on WebNLG and WikiBio are shown in Table \ref{tab:fewshotwebnlg} and \ref{tab:fewshotwikibio}. We can observe that CBST significantly outperforms fine-tuning and task-adaptive pre-training methods in all the settings, which shows that our method can explicitly learn the relationship between structured data and texts via self-training and improve the model performance. 
The comparison between CBST and the ablation models indicates that curriculum learning can alleviate the problem of low-quality pseudo-labeled data and further improve the performance. The noise functions also contribute to the final performance by increasing the smoothness of our model.

\subsection{Human Evaluation}
\label{sec:human}

\begin{table} [!t]
\centering
\scriptsize
\setlength{\tabcolsep}{1.0mm}{
\begin{tabular}{l|ccc|c}
\hline
Criterion & \multicolumn{3}{c|}{Fluency} & \multirow{2}*{$\kappa$}  \\
\cline{1-4}
Model & Win (\%) & Lose (\%) & Tie (\%) &  \\
\hline
CBST vs. FT-BART & 41.3* & 31.7 & 27.0 & 0.411   \\
\hline
CBST vs. TAPT-SeqRecon & 42.0** & 29.3  & 28.7 & 0.410   \\
CBST vs. TAPT-GraphRecon & 41.0* & 32.0 & 27.0 & 0.492   \\
\hline
CBST vs. CBST w/o CL & 40.3**  & 28.0 & 31.7 & 0.438  \\
CBST vs. CBST w/o CL \& Noise & 44.7** & 27.0 & 28.3 & 0.403 \\
\hline
\hline
Criterion & \multicolumn{3}{c|}{Adequacy} & \multirow{2}*{$\kappa$}  \\
\cline{1-4}
Model & Win (\%) & Lose (\%) & Tie (\%) &  \\
\hline
CBST vs. FT-BART & 58.7** & 25.0 & 16.3  & 0.424   \\
\hline
CBST vs. TAPT-SeqRecon & 48.0** & 26.3 & 25.7  & 0.427   \\
CBST vs. TAPT-GraphRecon & 44.0* & 32.7 & 23.3 & 0.401   \\
\hline
CBST vs. CBST w/o CL & 40.0* & 31.3 & 28.7 & 0.495  \\
CBST vs. CBST w/o CL \& Noise & 51.0** & 28.3 & 20.7 & 0.449 \\
\hline
\end{tabular}}
\caption{Human evaluation on different methods in the 1\% setting of WebNLG. The scores indicate the percentages of win, lose and tie when CBST is compared with other baselines. $\kappa$ is Fleiss' Kappa (all indicate moderate agreement). The scores marked with * mean $p<0.05$ while ** means $p<0.01$ in sign test.}
\label{tab:webnlghuman}
\end{table}

To further evaluate the quality of generated results, we conducted human evaluation in the 1\% setting of WebNLG.
We followed the existing works \cite{ribeiro2020globallocal} to use two criteria: \textit{fluency} (whether a sentence is grammatically fluent) and \textit{adequacy} (whether a sentence clearly describes the structured data). We randomly sampled 100 structured data from the test set, and collected the generated results from CBST and other baselines. We adopted pairwise comparison \cite{ke2021jointgt} between CBST and other baselines. For each pair of generated texts (one from CBST and the other from the corresponding baseline, given the same input structured data), three annotators were hired to determine which text is better (i.e., win, lose or tie) in terms of the above metrics.

Results in Table \ref{tab:webnlghuman} show that CBST can significantly outperform the baselines based on direct fine-tuning and task-adaptive pre-training in both fluency and adequacy.
In addition, the improvement of CBST over two ablation models shows the effectiveness of our curriculum learning module and noise functions to generate fluent texts
which describe structured data more clearly.
We also calculated Fleiss' Kappa \cite{fleiss1971kappa} for each pairwise comparison to measure the agreement among different annotators, where the results in Table \ref{tab:webnlghuman} show moderate agreement ($0.4\leq \kappa \leq 0.6$).

\subsection{Ablation Study}
\label{sec:ablation}
\begin{table} [!h]
\centering
\scriptsize
\setlength{\tabcolsep}{1.0mm}{
\begin{tabular}{l|ccc}
\hline
Model & B-4 & M & R-L  \\
\hline
CBST & \textbf{44.40} & \textbf{37.37} & \textbf{63.62} \\
\hline
\quad w/o NoiseTriple & 43.87 & 37.24 & 62.92 \\
\quad w/o NoiseWord & 43.53 & 36.93 & 62.94 \\
\quad w/o SelectCov & 43.74 & 37.08 & 63.37 \\
\quad w/o SelectProb & 43.53 & 37.19 & 62.44 \\
\hline
\quad w/ DiffLen & 43.93 & 36.75 & 63.27 \\
\hline
\end{tabular}}
\caption{Ablation test. NoiseTriple / NoiseWord / SelectCov / SelectProb / DiffLen denotes triple reordering / word substitution / coverage-based selection / probability-based selection / length-based difficulty metric, respectively.}
\label{tab:ablation}
\end{table}
We conducted a detailed ablation test to study the effects of different noise functions, selection criteria, and difficulty metrics.
We removed each module respectively and presented the results on WebNLG (1\%) in Table \ref{tab:ablation}. Note that \textit{w/ DiffLen} denotes that we used the length of linearized structured data as the difficulty metric.

\begin{table} [!t]
\centering
\scriptsize
\setlength{\tabcolsep}{1.0mm}{
\begin{tabular}{l|c|c|c}
\hline
Model & Hallucination & Missing Fact & Fluency  \\
\hline
CBST  & \textbf{11.3\%} & \textbf{19.0\%} & \textbf{4.45} \\
\quad w/o CL & 17.7\% & 19.3\% & 4.02 \\
\quad w/o CL \& Noise & 18.0\% & 19.7\% & 3.97 \\
\hline
\end{tabular}}
\caption{Human evaluation on the pseudo-labeled data generated by different methods at the last iteration.}
\label{tab:pseudodata}
\end{table}

Results in Table \ref{tab:ablation} show that all the modules contribute to the final performance.
As for two noise functions, the performance of CBST degrades more in the setting of removing word substitution, which perturbs structured data more flexibly.
In terms of the selection criterion, both coverage and generation probability improve the quality of pseudo-labeled data and contribute to the final performance. We can also observe the performance drop when CBST used the length of linearized structured data as the difficulty metric, indicating that the number of input triples is a more proper metric to reflect the difficulty of text generation from structured data.

\subsection{Analysis on Pseudo-Labeled Data}

To study whether our method can improve the quality of pseudo-labeled data via curriculum learning, we evaluated the quality of pseudo-labeled data generated by the teacher model at the last iteration before selection.
We resorted to human evaluation since there is no annotated text for unlabeled structured data. We randomly sampled 100 unlabeled structured data and collected the generated results of the teacher models from CBST and the ablation models at the last iteration. Three annotators were hired to judge each generated text from the following fine-grained aspects: 1) \textit{Hallucination}: whether the generated text includes non-existing facts; 2) \textit{Missing fact}: whether the generated text misses input facts; 3) \textit{Fluency}: the fluency score of generated texts (score 1-5 where 5 indicates fully fluent sentences).

We presented the fluency score and the proportions of generated results that belong to hallucination / missing fact in Table \ref{tab:pseudodata}. Results show that curriculum learning in our method plays an important role in generating fluent and adequate texts to describe unlabeled structured data, resulting in better model performance. The relatively limited improvement on missing facts may be because our base model BART already has the strong ability to generate texts that appear in the input via its pre-training tasks based on text reconstruction.

\subsection{Analysis on Curriculum Learning}
\label{sec:analysiscur}

To further analyze how curriculum learning helps self-training in few-shot data-to-text generation, we first demonstrated how the number of curriculum ($M_C$) affects the final performance. The results in Table \ref{tab:differentmc} show that the best performance is reached at $M_C=3$.
When $M_C$ is smaller, the teacher model needs to generate texts for unlabeled structured data with multiple triples at early iterations, which may degrade the quality of pseudo-labeled data and the final performance. In contrast, when $M_C$ is larger, the student model is trained on easy unlabeled data for many iterations and cannot utilize the hard cases until the late iterations, which may also affect the model performance.

\begin{table} [!t]
\centering
\scriptsize
\setlength{\tabcolsep}{1.0mm}{
\begin{tabular}{c|c|c|c}
\hline
$M_C$ & B-4 & M & R-L  \\
\hline
2 & 43.55 & 36.79 & 62.93  \\
3 & \textbf{44.40} & \textbf{37.37} & \textbf{63.62} \\
4 & 43.29 & 36.67 & 63.07   \\
\hline
\end{tabular}}
\caption{BLEU-4(B-4), METEOR(M), ROUGE-L(R-L) on WebNLG (1\%) with different numbers of curriculum.}
\label{tab:differentmc}
\end{table}

\begin{figure}[!t]
  \centering
  \includegraphics[width=1.0\linewidth]{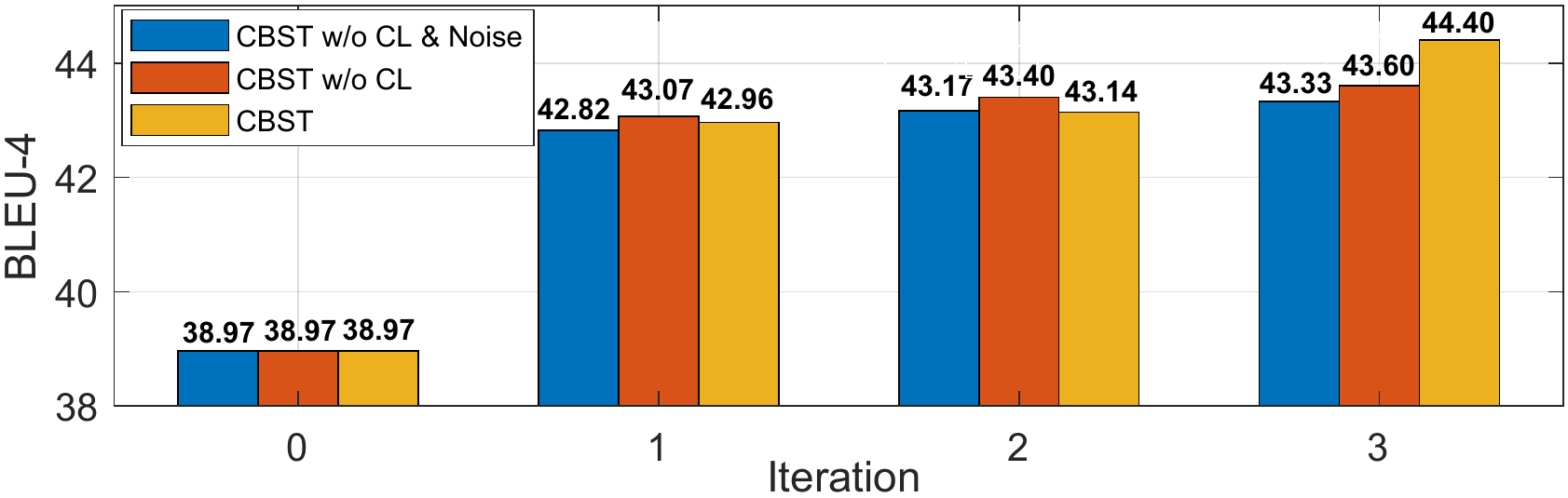}% 1\linewidth
  \caption{BLEU-4 of CBST and two ablation models at different iterations.}
  \label{fig:curriculum}
\end{figure}

Furthermore, we set $M_C=3$ and visualized the performance of the student model at each iteration in Figure \ref{fig:curriculum}. Note that the values when the number of iterations equals 0 indicate the performance of direct fine-tuning.
At early iterations, the two ablation models perform better because they directly incorporate the whole unlabeled dataset into self-training and acquire a larger pseudo-labeled dataset.
However, the improvement of their performance is extremely limited at the second and third iterations since the low-quality texts generated by the teacher model at early iterations may restrict the model performance. For comparison, CBST only utilizes the unlabeled data that satisfy the difficulty metric at each iteration to avoid low-quality pseudo-labeled data. Despite the worse performance at early iterations due to the smaller number of unlabeled data included in the self-training process, CBST still obtains the best performance at the last iteration.

\section{Conclusion}

We introduce self-training to improve text-to-text pre-trained models on few-shot data-to-text generation tasks, which can utilize unlabeled structured data to explicitly model the relationship between structured data and texts. To alleviate the problem of low-quality pseudo-labeled data during self-training, we further propose a novel method called Curriculum-Based Self-Training to rearrange the order of unlabeled data incorporated into self-training based on the difficulty metric. Experimental results show that CBST outperforms fine-tuning and task-adaptive pre-training methods, and achieves state-of-the-art performance in the few-shot setting of WebNLG and WikiBio datasets.

\section*{Acknowledgments}

This work was supported by the National Science Foundation for Distinguished Young Scholars (with No. 62125604) and the NSFC projects (Key project with No. 61936010 and regular project with No. 61876096). This work was supported by the Guoqiang Institute of Tsinghua University, with Grant No. 2019GQG1 and 2020GQG0005. This work was also supported by OPPO Research Fund. This work was sponsored by Tsinghua-Toyota Joint Research Fund.

\bibliographystyle{named}
\bibliography{ijcai22}

\end{document}